\documentclass[letterpaper]{article} 
\usepackage[preprint]{aaai2027}  
\usepackage[hyphens]{url}  
\usepackage{graphicx} 
\usepackage{natbib}  
\usepackage{caption} 

\usepackage[table]{xcolor}
\usepackage{colortbl}
\usepackage{booktabs}
\usepackage{multirow}
\usepackage{makecell}
\usepackage{graphicx}
\usepackage{adjustbox}
\usepackage{xspace}

\usepackage{algorithm}
\usepackage{algorithmic}
\usepackage{amsmath}
\usepackage{tabularx}
\usepackage{array}
\usepackage{pifont}
\usepackage{newfloat}
\usepackage{listings}
\DeclareCaptionStyle{ruled}{labelfont=normalfont,labelsep=colon,strut=off} 
\floatstyle{ruled}
\newfloat{listing}{tb}{lst}{}
\floatname{listing}{Listing}

\usepackage{booktabs}

\newcommand{\vnum}{\textit{VidNum}\xspace}
\title{\textit{VidNum}: Diagnosing VLM Failure Modes in Video-Grounded Numerical Reasoning}
\author{
    Shaoyang Cui\equalcontrib\textsuperscript{\rm 1},
    Lingbei Meng\equalcontrib\textsuperscript{\rm 2},
    Yaodi Luo\equalcontrib\textsuperscript{\rm 3},
    Peize He\equalcontrib\textsuperscript{\rm 3}
}
\affiliations{
    \textsuperscript{\rm 1}School of Psychological and Cognitive Sciences,
    Peking University, Beijing, China\\
    \textsuperscript{\rm 2}Shenzhen Loop Area Institute,
    Shenzhen, Guangdong, China\\
    \textsuperscript{\rm 3}University of Electronic Science and Technology of China,
    Chengdu, Sichuan, China\\
    sy-cui@pku.edu.cn
}

\definecolor{OpenColor}{HTML}{1F77B4}
\definecolor{ClosedColor}{HTML}{8E44AD}
\definecolor{TableHeadBG}{HTML}{EEF4FF}
\definecolor{OpenRowBG}{HTML}{F5FAFF}
\definecolor{ClosedRowBG}{HTML}{F8F1FF}
\definecolor{PromptBlueBG}{HTML}{F2F8FF}
\definecolor{PromptBlueFrame}{HTML}{4A90E2}
\definecolor{PromptOrangeBG}{HTML}{FFF7ED}
\definecolor{PromptOrangeFrame}{HTML}{F39C12}
\definecolor{MarkGreen}{HTML}{2E7D32}
\definecolor{MarkRed}{HTML}{C62828}
\definecolor{MarkGold}{HTML}{B7791F}
\newcommand{\cmark}{\textcolor{MarkGreen}{\ding{51}}}
\newcommand{\xmark}{\textcolor{MarkRed}{\ding{55}}}
\newcommand{\pmark}{\textcolor{MarkGold}{\ding{108}}}

\begin{document}

\maketitle

\begin{abstract}
Video-grounded numerical reasoning requires Vision-Language Models (VLMs) to
identify, track, and combine quantitative evidence across frames, actions, and
scene changes. Existing benchmarks provide fragmented coverage: general
VideoQA includes counting among broader tasks, while dedicated benchmarks
focus on repetition counting, ultra-long-video enumeration, or instructional
mathematics. We introduce \textbf{\vnum}, a manually curated and independently
verified benchmark containing 1,167 multiple-choice questions. Its three task
groups distinguish \textbf{Direct and Distinct Enumeration}, \textbf{Conditioned
and Structured Enumeration}, and \textbf{Compositional Quantitative Reasoning}.
Question-level annotations further identify the evidence target, counting
structure, and required reasoning operation. The best evaluated VLM reaches
59.8\% accuracy, compared with 98.2\% for human annotators, and no evaluated
open-weight model exceeds 45\%. Stratified analyses reveal that failures are
not uniformly distributed: structured target construction and
action-grounded compositional reasoning form recurring bottlenecks across
models. Zero-shot chain-of-thought prompting is not a reliable remedy: it
recovers some errors but breaks previously correct answers, with effects that
vary across models and task structures. \vnum therefore supports diagnostic
analysis beyond a single aggregate score.

\end{abstract}

\begin{table*}[t]
\centering
\caption{Comparison with representative counting and quantitative video benchmarks.}
\label{tab:counting-positioning}

\footnotesize
\renewcommand{\arraystretch}{1.25}
\setlength{\tabcolsep}{3.2pt}
\arrayrulecolor{OpenColor!45}

\begin{tabular}{
>{\raggedright\arraybackslash}p{0.215\textwidth}
>{\centering\arraybackslash}p{0.105\textwidth}
>{\centering\arraybackslash}p{0.08\textwidth}
>{\centering\arraybackslash}p{0.10\textwidth}
>{\centering\arraybackslash}p{0.10\textwidth}
>{\centering\arraybackslash}p{0.10\textwidth}
>{\raggedright\arraybackslash}p{0.205\textwidth}}
\toprule
\rowcolor{TableHeadBG}
\textbf{Benchmark} &
\textbf{Scale} &
\makecell[c]{\small\textbf{Target}\\\small\textbf{diversity}} &
\makecell[c]{\small\textbf{Conditioned}\\\small\textbf{targets}} &
\makecell[c]{\small\textbf{Quantity}\\\small\textbf{composition}} &
\makecell[c]{\small\textbf{Item-level}\\\small\textbf{labels}} &
\textbf{Primary setting} \\
\midrule

\rowcolor{OpenColor!10}
\multicolumn{7}{l}{\textbf{General video question answering}} \\
\mbox{TGIF-QA~\citeyearpar{jang2017tgifqa}} & \mbox{30,397 QA} & \xmark{} & \xmark{} & \xmark{} & \xmark{} & \mbox{\scriptsize Repetition counting} \\
\mbox{MVBench~\citeyearpar{li2024mvbench}} & \mbox{400 MCQ} & \cmark{} & \xmark{} & \xmark{} & \pmark{} & \mbox{\scriptsize General VideoQA} \\

\midrule
\rowcolor{OpenColor!10}
\multicolumn{7}{l}{\textbf{Counting and quantitative video benchmarks}} \\
\mbox{Countix~\citeyearpar{dwibedi2020countingouttime}} & \mbox{8,757 videos} & \xmark{} & \xmark{} & \xmark{} & \xmark{} & \mbox{\scriptsize Cycle counting} \\
\mbox{UCFRep~\citeyearpar{zhang2020contextaware}} & \mbox{526 videos} & \xmark{} & \xmark{} & \xmark{} & \xmark{} & \mbox{\scriptsize Cycle counting} \\
\mbox{OVR~\citeyearpar{dwibedi2024ovr}} & \mbox{$>$72K ann.} & \xmark{} & \xmark{} & \xmark{} & \pmark{} & \mbox{\scriptsize Open-vocabulary repetition} \\
\mbox{CG-AV-Counting~\citeyearpar{lu2025avreasoner}} & \mbox{1,027 QA} & \cmark{} & \cmark{} & \xmark{} & \cmark{} & \mbox{\scriptsize Clue-grounded AV counting} \\
\mbox{SVCBench~\citeyearpar{liu2026svcbench}} & \mbox{1,000 QA} & \cmark{} & \pmark{} & \xmark{} & \cmark{} & \mbox{\scriptsize Streaming state maintenance} \\
\mbox{EC-Bench~\citeyearpar{tsuchiya2026ecbench}} & \mbox{1,699 queries} & \cmark{} & \cmark{} & \xmark{} & \cmark{} & \mbox{\scriptsize Ultra-long enumeration} \\
\mbox{VideoMathQA~\citeyearpar{rasheed2026videomathqa}} & \mbox{420 MCQ} & \xmark{} & \xmark{} & \cmark{} & \cmark{} & \mbox{\scriptsize Instructional math} \\

\midrule
\rowcolor{OpenColor!16}
\mbox{\textbf{VidNum (ours)}} & \mbox{\textbf{1,167 MCQ}} & \cmark{} & \cmark{} & \cmark{} & \cmark{} &
\mbox{\scriptsize\textbf{Diagnostic numerical VideoQA}} \\
\bottomrule
\end{tabular}

\vspace{3pt}
\scriptsize
\textit{Notes.}
\cmark{} denotes explicit benchmark-level support, \pmark{} partial or adjacent
support, and \xmark{} no explicit support. ``Multiple target types'' requires
more than one family of countable visual evidence. ``Conditioned targets''
denotes target selection by attributes, relations, temporal scope, or event
structure. ``Quantity composition'' requires an operation over video-derived
quantities. ``Item-level diagnostic labels'' denotes annotations beyond the
final answer. TGIF-QA reports the \emph{Repetition Count} split; MVBench combines
its \texttt{action\_count} and \texttt{moving\_count} subsets.
\arrayrulecolor{black}
\end{table*}

\section{Introduction}

\begin{figure*}[h]
  \centering
  \includegraphics[width=\textwidth]{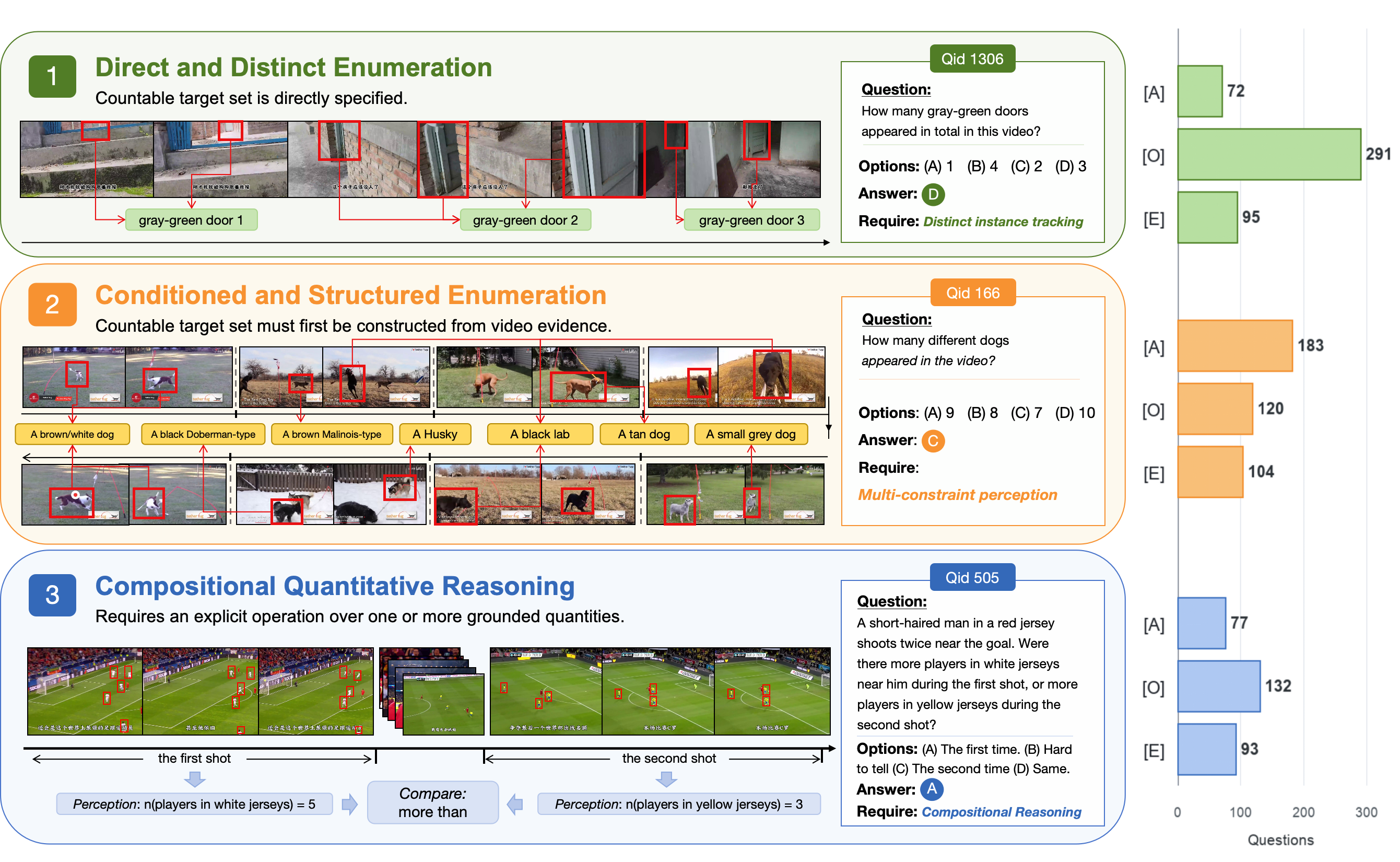}
  \caption{
    Overview of \vnum. The benchmark covers three diagnostic task groups:
    Direct and Distinct Enumeration (DDE), Conditioned and Structured Enumeration
    (CSE), and Compositional Quantitative Reasoning (CQR). Examples illustrate
    different requirements for target specification, visual grounding, and quantity
    composition. The right panel shows the question distribution by target type,
    where [A], [O], and [E] denote action, object, and event targets.
    }
  \label{fig:teaser}
  
\end{figure*}

Recent vision-language models (VLMs) have substantially advanced video understanding, enabling question answering over increasingly long and diverse videos. Large-scale benchmarks now evaluate temporal perception, event understanding, and multimodal reasoning across a wide range of scenarios \cite{li2024mvbench,fu2025videomme}. Within this broad landscape, video-grounded numerical reasoning represents a particularly informative capability. To answer a numerical question, a model must transform observations distributed across frames, actions, and scene changes into explicit quantities, such as how many entities appear, how often an action occurs, or how two event counts compare. This process culminates in a discrete, objectively verifiable answer, yet still depends on accurate spatial, temporal, and relational grounding (Figure~\ref{fig:teaser}).

This combination gives video-grounded numerical reasoning a distinctive diagnostic property. An incorrect answer can originate from multiple stages of the reasoning process: the model may construct the wrong target set, merge or duplicate identities, bind conditions to incorrect entities, segment actions improperly, or manipulate correctly grounded quantities incorrectly. Unlike many open-ended VideoQA tasks, numerical questions expose these failures through exact final answers while still requiring complex visual reasoning. Consequently, they provide a natural setting for analyzing how numerical reasoning fails, rather than only measuring whether a question is answered correctly.

Existing benchmarks, however, are not designed around this diagnostic perspective. General VideoQA datasets such as TGIF-QA, NExT-QA, and ActivityNet-QA combine quantitative questions with broader temporal and semantic understanding tasks \cite{jang2017tgifqa,xiao2021nextqa,yu2019activitynetqa}. More recent benchmarks, including MVBench and Video-MME, substantially broaden temporal and domain coverage, but numerical reasoning remains only one capability among many \cite{li2024mvbench,fu2025videomme}. Dedicated counting benchmarks instead focus on specific settings such as repeated-action counting \cite{dwibedi2020countingouttime,zhang2020contextaware}, clue-grounded or streaming counting \cite{lu2025avreasoner,liu2026svcbench}, ultra-long-video enumeration \cite{tsuchiya2026ecbench}, or mathematical reasoning in instructional videos \cite{rasheed2026videomathqa}. 
Together, these resources significantly improve task coverage, yet they do not
provide a unified question-level diagnostic taxonomy for comparing where
video-grounded numerical reasoning fails across target construction, evidence
grounding, and quantity composition.

The missing resource is therefore not simply another collection of counting questions, but a benchmark explicitly designed for diagnostic analysis. Such a benchmark should jointly characterize what visual evidence is quantified, how the countable target is constructed, and whether the resulting quantities must be further manipulated. These properties must also be annotated at the question level, allowing failures to be systematically compared across models instead of being summarized by aggregate accuracy alone.

To address this gap, we introduce \textbf{\vnum}, a manually curated and independently verified benchmark containing \textbf{1,167 multiple-choice questions} for video-grounded numerical reasoning. Rather than organizing questions by semantic topic, \vnum derives its taxonomy from the computation required to answer each question. Direct and Distinct Enumeration (DDE) covers directly specified instances and unique categories. Conditioned and Structured Enumeration (CSE) requires constructing the target set from attributes, relations, temporal segments, or event structure. Compositional Quantitative Reasoning (CQR) requires explicit operations over one or more grounded quantities. Orthogonal annotations further identify whether the evidence concerns actions, objects, or events, together with finer counting-structure and reasoning-operation labels. Consequently, the benchmark supports analysis beyond aggregate counting accuracy by enabling performance to be stratified according to target construction and quantitative reasoning requirements. Table~\ref{tab:counting-positioning} positions this design using benchmark-independent properties rather than the names of our task groups.

Experiments on representative open-weight and proprietary VLMs reveal three recurring diagnostic patterns. Structured target construction remains a bottleneck across models; action-grounded compositional reasoning emerges as a recurring cross-model failure pattern; and zero-shot chain-of-thought prompting redistributes item-level correctness rather than providing reliable gains. These findings demonstrate how \vnum transforms aggregate benchmark evaluation into a structured diagnosis of numerical reasoning failures.

Our contributions are threefold:
\begin{itemize}
\item We introduce \textbf{\vnum}, a manually curated and independently verified diagnostic benchmark of 1,167 questions spanning diverse visual targets and quantitative structures.
\item We derive a computation-oriented taxonomy from target-set construction and quantity use, together with question-level evidence-target, counting-structure, and reasoning-operation annotations that support controlled diagnostic analysis.
\item We evaluate representative VLMs and identify structured, cross-model failure patterns that would be obscured by aggregate accuracy alone.
\end{itemize}

\section{Related Work}

\subsection{VideoQA and Video-MLLM Evaluation}

VideoQA benchmarks have long evaluated a broad range of video understanding
abilities, including temporal reasoning, causal inference, event recognition,
and question answering over everyday activities. Early datasets such as
TGIF-QA~\cite{jang2017tgifqa}, ActivityNet-QA~\cite{yu2019activitynetqa}, and
NExT-QA~\cite{xiao2021nextqa} established video question answering as a useful
setting for studying temporal and causal understanding. More recent diagnostic
benchmarks have expanded the scope of evaluation. The Perception
Test~\cite{patraucean2023perception} evaluates perception and reasoning skills
through densely annotated real-world videos, while MVBench~\cite{li2024mvbench}
and Video-MME~\cite{fu2025videomme} assess a broad set of temporal abilities
across diverse tasks and video durations.

These benchmarks are essential for measuring general video understanding.
However, quantitative questions are typically one capability among many, and
their evaluation does not isolate the distinct processes involved in forming a
countable target set, maintaining identities across time, aggregating events,
or combining video-grounded quantities. \vnum complements these benchmarks by explicitly organizing numerical reasoning according to the computation required to answer each question.

\subsection{Video Counting and Quantification}

A separate line of work studies temporal repetition counting. QUVA
Repetition~\cite{runia2018realworld}, Countix and
RepNet~\cite{dwibedi2020countingouttime}, and
UCFRep~\cite{zhang2020contextaware} estimate the number of repeated action
cycles from video. These tasks have driven progress in periodicity modeling,
temporal self-similarity, and action-boundary estimation, but they are commonly
formulated as regression problems over repeated motions rather than
language-grounded VideoQA. More recent benchmarks provide richer diagnostic
signals. CG-AV-Counting~\cite{lu2025avreasoner} pairs long audiovisual videos
with fine-grained counting clues, while SVCBench~\cite{liu2026svcbench} uses
streaming query points and object/event subcategories to diagnose temporal
state maintenance. EC-Bench~\cite{tsuchiya2026ecbench} evaluates enumeration,
counting, and evidence grounding in videos longer than 30 minutes using
explicit temporal spans. These resources diagnose clue localization,
streaming state updates, and long-context enumeration. \vnum instead organizes numerical reasoning according to how countable targets are constructed and how grounded quantities are subsequently used.

\subsection{Visual and Video Numerical Reasoning}

Visual numerical reasoning has also been studied in images. CLEVR provides
controlled compositional reasoning over synthetic scenes~\cite{johnson2017clevr},
whereas TallyQA examines complex open-ended counting questions involving
attributes and relations in natural images~\cite{acharya2019tallyqa}. These
datasets demonstrate that counting depends on more than object detection, but
they do not require temporal tracking or event segmentation. VideoMathQA~\cite{rasheed2026videomathqa} extends mathematical reasoning to
educational videos, where models must integrate visual content, narration, and
textual information distributed over time. In contrast, \vnum focuses on video-grounded numerical reasoning across diverse visual targets and organizes questions according to their underlying computational requirements. Question-level annotations further enable fine-grained analysis across task groups, evidence targets, and reasoning operations.

\section{The \vnum}
\label{sec:dataset}

\vnum is a manually curated and independently verified benchmark for
video-grounded numerical reasoning. It contains 1,167 four-option
multiple-choice questions derived from 947 source videos. A single source
video may support multiple questions when different temporal segments or
distinct quantitative phenomena are examined. Each question consists of a
video segment, a natural-language question, four answer options, and a
verified answer. Detailed statistics on topic coverage and video duration are
provided in the technical appendix.

\begin{figure}[t]
  \centering
  \includegraphics[width=\linewidth]{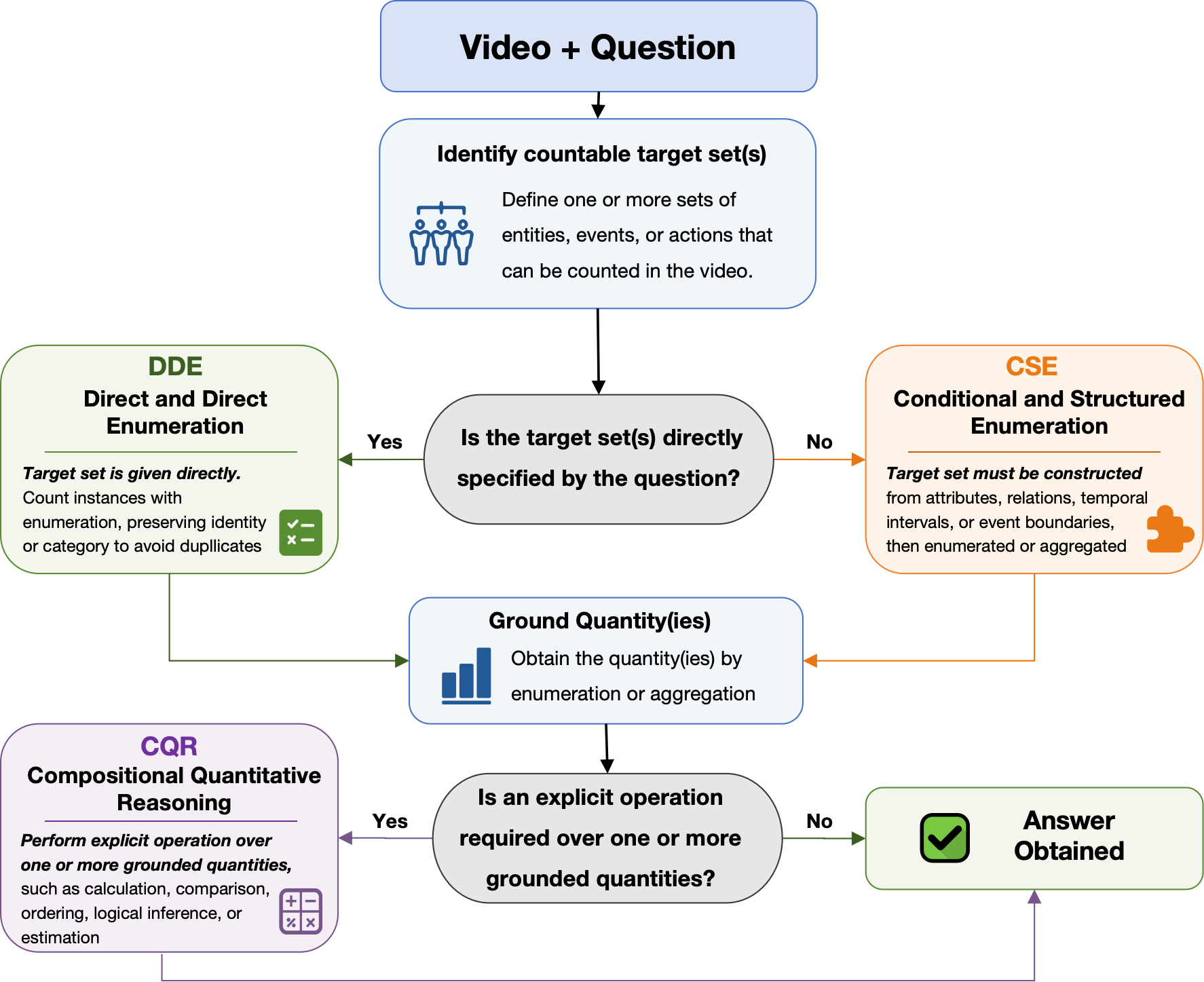}
  \caption{Computation-oriented overview of the \vnum task taxonomy.}
  \label{fig:taxonomy}
\end{figure}

Figure~\ref{fig:taxonomy} summarizes the computation-oriented design of the
proposed taxonomy. The taxonomy is derived from the computational process
required to answer a video-grounded numerical question, organizing questions
according to the reasoning needed to obtain the final answer. Specifically, it
distinguishes three forms of computation: direct enumeration of an explicitly
specified target, construction of the target set from video evidence, and
quantitative composition over grounded quantities.

\paragraph{Direct and Distinct Enumeration (DDE).}
DDE covers questions whose countable target set is directly specified. The
model enumerates the corresponding instances while preserving entity identity
or category distinctions when necessary. The answer is obtained directly from
the grounded count without requiring additional quantitative operations.

\paragraph{Conditioned and Structured Enumeration (CSE).}
CSE covers questions whose target set must first be constructed from video
evidence. Construction may depend on attributes, spatial relations, temporal
intervals, event boundaries, or aggregation rules. The defining challenge is
therefore selecting and structuring the correct evidence before enumeration.

\paragraph{Compositional Quantitative Reasoning (CQR).}
CQR covers questions whose final answer requires explicit operations over one
or more grounded quantities. Such operations may include arithmetic,
comparison, ordering, logical inference, estimation, or combinations thereof.
Rather than replacing grounding, CQR builds upon the underlying target
construction and enumeration process.

In addition to the task taxonomy, each question is assigned one primary
evidence target according to the information that must be quantified. The
evidence target identifies the source of the quantitative evidence, providing
an annotation axis orthogonal to the task taxonomy and enabling performance to
be analyzed across different forms of visual grounding.

We distinguish three evidence targets. Object-centric questions require
quantifying visible entities or semantic categories. Action-centric questions
require identifying and counting occurrences of actions. Event-centric
questions require locating and quantifying temporally bounded events or state
changes. These categories characterize where the required quantity originates,
rather than different levels of reasoning difficulty. The benchmark contains
543 object-centric questions, 332 action-centric questions, and 292
event-centric questions (Figure~\ref{fig:teaser}).

\newcolumntype{L}[1]{>{\raggedright\arraybackslash}p{#1}}
\newcolumntype{C}[1]{>{\centering\arraybackslash}p{#1}}
\newcolumntype{Y}{>{\raggedright\arraybackslash}X}

\newcommand{\paircell}[2]{\shortstack{#1/#2}}
\newcommand{\bestopen}[1]{\textbf{\underline{#1}}}

\begin{table*}[t]
  \caption{Main results on \vnum under the diagnostic task taxonomy.
  Model results report \textit{NoCoT/CoT} accuracy (\%), while human results
  report a single accuracy value. Within each ownership block,
  the best result in each column is highlighted; open-weight maxima are
  additionally underlined.}
  \label{tab:vidnum-detailed-results}
  \centering
  \small
  \renewcommand{\arraystretch}{1.25}
  \setlength{\tabcolsep}{3.2pt}
  \arrayrulecolor{OpenColor!45}
  \begin{adjustbox}{max width=\textwidth}
    \begin{tabular}{l|ccc|ccc|ccc|c}
      \toprule
      \rowcolor{TableHeadBG}
      & \multicolumn{3}{c|}{\textbf{DDE}}
      & \multicolumn{3}{c|}{\textbf{CSE}}
      & \multicolumn{3}{c|}{\textbf{CQR}}
      & \textbf{Overall} \\
      \rowcolor{TableHeadBG}
      \multirow{-2}{*}{\textbf{Model}}
      & \textbf{Action} & \textbf{Object} & \textbf{Event}
      & \textbf{Action} & \textbf{Object} & \textbf{Event}
      & \textbf{Action} & \textbf{Object} & \textbf{Event}
      & \textbf{Avg.} \\
      \midrule
      \rowcolor{OpenColor!12}
      \multicolumn{11}{l}{\textbf{Open-weight models}} \\
      LLaVA-NeXT-7B & \paircell{25.00}{18.06} & \paircell{26.12}{19.93} & \paircell{21.05}{21.05} & \paircell{21.31}{16.39} & \paircell{26.67}{20.83} & \paircell{21.15}{24.04} & \paircell{23.38}{15.58} & \paircell{25.19}{23.66} & \paircell{16.13}{15.05} & \paircell{23.41}{19.55} \\
      InternVL2.5-1B & \paircell{38.89}{18.06} & \paircell{27.49}{17.87} & \paircell{29.47}{21.05} & \paircell{34.97}{12.02} & \paircell{28.33}{16.67} & \paircell{25.96}{13.46} & \paircell{27.27}{14.29} & \paircell{33.33}{15.91} & \paircell{23.66}{10.75} & \paircell{29.82}{15.68} \\
      InternVL2.5-2B & \paircell{31.94}{18.06} & \paircell{30.24}{21.99} & \paircell{47.37}{29.47} & \paircell{26.78}{20.77} & \paircell{21.67}{23.33} & \paircell{30.77}{16.35} & \paircell{25.97}{22.08} & \paircell{36.36}{21.21} & \paircell{31.18}{12.90} & \paircell{30.85}{20.99} \\
      InternVL2.5-4B & \paircell{34.72}{36.11} & \paircell{34.36}{33.33} & \paircell{40.00}{35.79} & \paircell{28.96}{26.23} & \paircell{34.17}{33.33} & \paircell{26.92}{26.92} & \paircell{36.36}{35.06} & \paircell{37.12}{33.33} & \paircell{35.48}{29.03} & \paircell{33.85}{31.79} \\
      InternVL2.5-8B & \paircell{38.89}{40.28} & \paircell{35.05}{39.86} & \paircell{42.11}{41.05} & \paircell{31.15}{32.79} & \paircell{35.83}{34.17} & \paircell{30.77}{37.50} & \paircell{35.06}{33.77} & \paircell{42.75}{48.85} & \paircell{35.48}{44.09} & \paircell{35.85}{39.02} \\
      InternVL3-1B & \paircell{31.94}{30.56} & \paircell{26.46}{28.18} & \paircell{33.68}{27.37} & \paircell{30.05}{28.42} & \paircell{26.67}{23.33} & \paircell{29.81}{29.81} & \paircell{36.36}{18.18} & \paircell{34.09}{26.52} & \paircell{33.33}{22.58} & \paircell{30.33}{26.65} \\
      InternVL3-2B & \paircell{30.56}{33.33} & \paircell{30.93}{28.52} & \paircell{43.16}{31.58} & \paircell{31.15}{24.04} & \paircell{32.50}{32.50} & \paircell{35.58}{25.96} & \paircell{25.97}{25.97} & \paircell{37.12}{39.39} & \paircell{34.41}{29.03} & \paircell{33.16}{29.65} \\
      InternVL3-8B & \paircell{45.83}{\bestopen{48.61}} & \paircell{37.80}{40.21} & \paircell{45.26}{46.32} & \paircell{33.33}{32.79} & \paircell{30.00}{35.00} & \paircell{33.65}{39.42} & \paircell{33.77}{31.17} & \paircell{40.46}{48.85} & \paircell{44.09}{41.94} & \paircell{37.56}{39.97} \\
      InternVL3-14B & \paircell{37.50}{43.06} & \paircell{41.92}{42.96} & \paircell{\bestopen{48.42}}{50.53} & \paircell{28.42}{29.51} & \paircell{30.00}{\bestopen{41.67}} & \paircell{33.65}{35.58} & \paircell{32.47}{40.26} & \paircell{42.75}{54.20} & \paircell{\bestopen{50.54}}{48.39} & \paircell{38.25}{42.20} \\
      InternVL3-38B & \paircell{45.83}{38.89} & \paircell{41.92}{42.96} & \paircell{\bestopen{48.42}}{\bestopen{54.74}} & \paircell{30.05}{32.24} & \paircell{41.67}{39.17} & \paircell{\bestopen{42.31}}{\bestopen{41.35}} & \paircell{36.36}{38.96} & \paircell{\bestopen{53.44}}{56.49} & \paircell{46.24}{40.86} & \paircell{\bestopen{42.11}}{42.54} \\
      InternVL3-78B & \paircell{38.89}{44.44} & \paircell{\bestopen{42.61}}{\bestopen{49.83}} & \paircell{47.37}{51.58} & \paircell{28.42}{31.69} & \paircell{\bestopen{49.17}}{37.50} & \paircell{36.54}{33.65} & \paircell{37.66}{38.96} & \paircell{50.38}{\bestopen{58.02}} & \paircell{44.09}{\bestopen{58.06}} & \paircell{41.34}{\bestopen{44.94}} \\
      InternVL3.5-1B & \paircell{18.06}{30.56} & \paircell{27.49}{27.49} & \paircell{30.53}{26.32} & \paircell{30.05}{\bestopen{33.88}} & \paircell{25.00}{25.83} & \paircell{28.85}{25.00} & \paircell{23.38}{\bestopen{41.56}} & \paircell{28.03}{31.82} & \paircell{23.66}{31.18} & \paircell{26.91}{29.91} \\
      InternVL3.5-2B & \paircell{37.50}{22.22} & \paircell{30.93}{30.24} & \paircell{43.16}{44.21} & \paircell{28.42}{21.31} & \paircell{34.17}{24.17} & \paircell{33.65}{14.42} & \paircell{32.47}{18.18} & \paircell{40.91}{25.00} & \paircell{37.63}{30.11} & \paircell{34.28}{26.05} \\
      InternVL3.5-4B & \paircell{45.83}{13.89} & \paircell{39.18}{20.96} & \paircell{44.21}{14.74} & \paircell{\bestopen{37.16}}{9.29} & \paircell{28.33}{15.00} & \paircell{30.77}{12.50} & \paircell{40.26}{11.69} & \paircell{40.15}{19.70} & \paircell{36.56}{18.28} & \paircell{37.79}{15.85} \\
      InternVL3.5-8B & \paircell{38.89}{36.11} & \paircell{34.71}{37.11} & \paircell{36.84}{45.26} & \paircell{33.33}{31.15} & \paircell{32.50}{37.50} & \paircell{29.81}{35.58} & \paircell{35.06}{37.66} & \paircell{48.09}{44.27} & \paircell{38.71}{48.39} & \paircell{36.11}{38.42} \\
      Qwen2.5-VL-3B-Instruct & \paircell{\bestopen{50.00}}{43.06} & \paircell{31.27}{30.24} & \paircell{36.84}{35.79} & \paircell{34.97}{30.60} & \paircell{30.00}{31.67} & \paircell{33.65}{26.92} & \paircell{32.47}{28.57} & \paircell{41.67}{28.79} & \paircell{30.11}{34.41} & \paircell{34.70}{31.45} \\
      Qwen2.5-VL-7B-Instruct & \paircell{33.33}{37.50} & \paircell{32.99}{32.99} & \paircell{43.16}{34.74} & \paircell{33.88}{25.68} & \paircell{32.50}{33.33} & \paircell{33.65}{23.08} & \paircell{38.96}{25.97} & \paircell{38.93}{38.93} & \paircell{36.56}{30.11} & \paircell{35.33}{31.39} \\
      Qwen3-VL-2B-Instruct & \paircell{34.72}{6.94} & \paircell{28.87}{14.78} & \paircell{35.79}{20.00} & \paircell{34.43}{12.02} & \paircell{28.33}{16.67} & \paircell{30.77}{9.62} & \paircell{35.06}{11.69} & \paircell{33.33}{18.18} & \paircell{33.33}{17.20} & \paircell{32.05}{14.40} \\
      Qwen3-VL-4B-Instruct & \paircell{38.89}{37.50} & \paircell{34.02}{37.46} & \paircell{40.00}{37.89} & \paircell{36.07}{23.50} & \paircell{21.67}{30.00} & \paircell{32.69}{31.73} & \paircell{32.47}{29.87} & \paircell{43.94}{33.33} & \paircell{35.48}{37.63} & \paircell{34.88}{33.08} \\
      Qwen3-VL-8B-Instruct & \paircell{38.89}{27.78} & \paircell{25.77}{27.49} & \paircell{36.84}{31.58} & \paircell{28.96}{24.04} & \paircell{29.17}{20.83} & \paircell{34.62}{29.81} & \paircell{\bestopen{41.56}}{28.57} & \paircell{37.40}{35.11} & \paircell{36.56}{29.03} & \paircell{32.33}{27.87} \\
      Qwen3-VL-2B-Thinking & \paircell{20.83}{16.67} & \paircell{26.80}{28.18} & \paircell{25.26}{21.05} & \paircell{21.31}{25.68} & \paircell{11.67}{20.83} & \paircell{14.42}{25.96} & \paircell{10.39}{19.48} & \paircell{24.24}{24.24} & \paircell{20.43}{22.58} & \paircell{20.91}{24.08} \\
      Qwen3-VL-4B-Thinking & \paircell{27.78}{22.22} & \paircell{32.65}{22.68} & \paircell{34.74}{14.74} & \paircell{26.78}{16.94} & \paircell{26.67}{15.83} & \paircell{23.08}{21.15} & \paircell{27.27}{14.29} & \paircell{27.27}{20.45} & \paircell{33.33}{26.88} & \paircell{29.22}{19.79} \\
      \midrule
      \rowcolor{ClosedColor!12}
      \multicolumn{11}{l}{\textbf{Proprietary models}} \\
      Gemini-3-Flash & \paircell{48.61}{\textbf{47.22}} & \paircell{55.67}{54.64} & \paircell{58.95}{60.00} & \paircell{39.34}{37.70} & \paircell{50.00}{55.00} & \paircell{53.85}{55.77} & \paircell{44.16}{40.26} & \paircell{58.78}{61.83} & \paircell{75.27}{63.44} & \paircell{53.34}{52.66} \\
      Gemini-3.1-Pro & \paircell{\textbf{50.00}}{45.83} & \paircell{59.11}{57.39} & \paircell{\textbf{66.32}}{\textbf{67.37}} & \paircell{38.80}{44.81} & \paircell{45.00}{55.00} & \paircell{56.73}{62.50} & \paircell{46.75}{\textbf{45.45}} & \paircell{61.07}{\textbf{67.18}} & \paircell{70.97}{\textbf{77.42}} & \paircell{54.63}{57.63} \\
      GPT-5.5 & \paircell{\textbf{50.00}}{40.28} & \paircell{\textbf{64.60}}{\textbf{63.92}} & \paircell{65.26}{63.16} & \paircell{\textbf{40.44}}{\textbf{47.54}} & \paircell{\textbf{59.17}}{\textbf{62.50}} & \paircell{\textbf{65.38}}{\textbf{67.31}} & \paircell{\textbf{51.95}}{\textbf{45.45}} & \paircell{\textbf{64.39}}{63.64} & \paircell{\textbf{76.34}}{\textbf{77.42}} & \paircell{\textbf{59.55}}{\textbf{59.81}} \\
      \midrule
      \rowcolor{MarkGold!16}
      \multicolumn{11}{l}{\textbf{Human performance}} \\
      \rowcolor{MarkGold!8}
      \textbf{Human} & \textbf{100.00} & \textbf{97.25} & \textbf{97.89} & \textbf{98.91} & \textbf{95.83} & \textbf{100.00} & \textbf{98.70} & \textbf{98.48} & \textbf{98.92} & \textbf{98.20} \\
      \bottomrule
    \end{tabular}
  \end{adjustbox}
  \arrayrulecolor{black}
\end{table*}

\paragraph{Fine-Grained Annotations.}
To facilitate more detailed diagnostic analyses, \vnum further provides two
auxiliary annotation dimensions: counting structure and reasoning operation.
Counting-structure annotations distinguish Direct Homogeneous Enumeration,
Distinct Entity or Category Enumeration, Attribute- or Relation-Conditioned
Enumeration, Multi-Instance or Event Aggregation, and Compositional
Quantitative Reasoning. Reasoning-operation annotations specify whether the
final answer requires no explicit operation, calculation, comparison, logical
inference, estimation, or combinations thereof. Together with the task
taxonomy and evidence targets, these auxiliary annotations enable analyses at
multiple levels of granularity.

\subsection{Benchmark Reliability and Validation}

\vnum was constructed through independent question creation, primary
validation, quality review, and final audit. Ambiguous questions, inaccurate
timestamps, label errors, and examples answerable from language priors alone
were removed through this multi-stage verification process.

To assess annotation consistency, we independently re-annotated a
natural-distribution sample of 151 finalized questions using the DDE/CSE/CQR
definitions. Agreement with the released labels averaged \textbf{98.23\%};
all three annotators agreed on \textbf{96.69\%} of the sample, with
Fleiss' $\kappa=0.970$.

We further perform benchmark validity analyses to examine whether performance
reflects video-grounded numerical reasoning rather than artifacts of the
evaluation format. Specifically, we evaluate potential multiple-choice bias
through direct-answer evaluation and measure language-only shortcuts through
text-only evaluation. Additional construction details and validation analyses
are provided in the technical appendix.

\begin{figure*}[t]
  \centering
  \includegraphics[width=\textwidth]{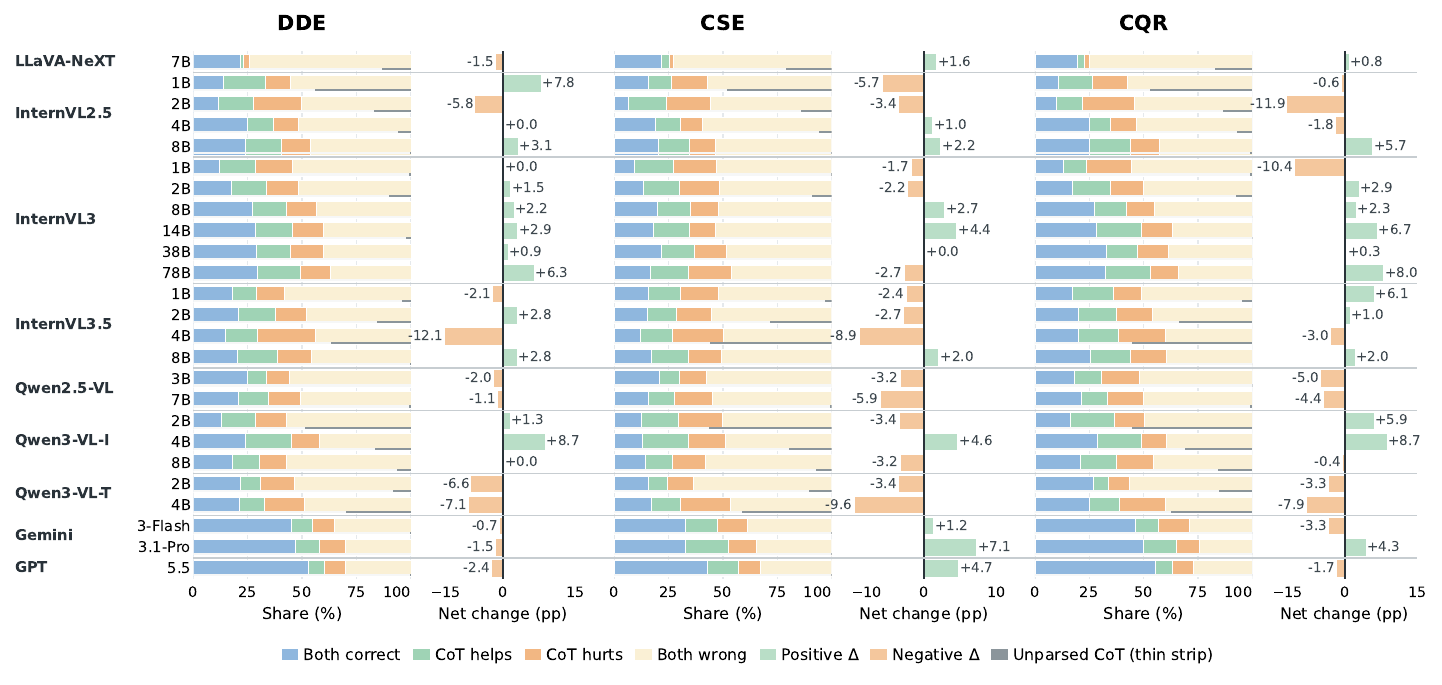}
  \caption{Item-level prediction transitions induced by zero-shot CoT across
    \vnum task groups. Each stacked bar is computed on questions with parsable
    NoCoT and CoT outputs. Green indicates items recovered by CoT, orange indicates
    previously correct items broken by CoT, and the right panel reports the net
    change, recovered minus broken, in percentage points. Thin gray strips show the end-to-end
    fraction of CoT outputs that could not be parsed. Models are grouped by family
    and ordered by scale within each family.}
    \label{fig:cot-transition-groups}
\end{figure*}

\section{Experiments}

\subsection{Evaluation Setup}

We evaluate 25 representative open-weight and proprietary VLMs. The suite
includes LLaVA-NeXT~\cite{liu2024llavanext}, InternVL2.5/3/3.5
\cite{chen2024internvl25,zhu2025internvl3,wang2025internvl35}, Qwen2.5-VL and
Qwen3-VL~\cite{bai2025qwen25vl,qwen2025qwen3vl}, two Gemini variants
\cite{google2025gemini3,google2026gemini31pro} and GPT-5.5. Each model is
tested with a NoCoT prompt that requests only the option letter and a zero-shot
CoT prompt that requests intermediate observations before the final option.
Both protocols require the choice inside \texttt{<answer>} tags. Unless noted
otherwise, we uniformly sample 48 frames from each video and preserve temporal
order. Main-table runs use the Chinese question and option fields. Invalid,
empty, or unparsable outputs are counted as incorrect in end-to-end accuracy;
the paired CoT analysis reports formatting failures separately. Full prompts,
answer-extraction rules, and input details are provided in the technical
appendix.

We evaluate 25 representative open-weight and proprietary VLMs. The evaluation
suite covers representative open-weight model families, including LLaVA-NeXT
\cite{liu2024llavanext}, InternVL2.5, InternVL3, and InternVL3.5
\cite{chen2024internvl25,zhu2025internvl3,wang2025internvl35}, Qwen2.5-VL and
Qwen3-VL~\cite{bai2025qwen25vl,qwen2025qwen3vl}, as well as proprietary models
including two Gemini variants
\cite{google2025gemini3,google2026gemini31pro} and GPT-5.5.
Each model is evaluated with a NoCoT prompt that requests only the final option
letter and a zero-shot CoT prompt that requests intermediate observations,
counting, or calculation steps before the final option. Both prompts require
the final choice to be provided inside \texttt{<answer>} tags. Unless noted
otherwise, we uniformly sample 48 frames from each video while preserving
temporal order. Main-table runs use the original Chinese question and option
fields to preserve the benchmark format. Invalid, empty, or unparsable outputs
are counted as incorrect in the end-to-end evaluation, while paired CoT
analyses separately report formatting failures. Full prompts, answer
extraction rules, and input details are provided in the technical appendix.

\subsection{Main Results}

Table~\ref{tab:vidnum-detailed-results} reports accuracy on \vnum under
\textit{NoCoT/CoT} evaluation. Current VLMs still exhibit substantial
difficulty on video-grounded numerical reasoning. The best evaluated model
achieves 59.81\% accuracy, compared with 98.20\% human performance, leaving a
large gap even for strong proprietary systems. Among open-weight models, the
best performance remains 44.94\%, indicating that accurately grounding and
reasoning over quantities in dynamic videos remains challenging for current
models.

The results also show clear differences across model families and task
structures. Larger and stronger models generally achieve better overall
performance, but improvements are not uniform across all categories. In
particular, the stratified results reveal substantial variation across
enumeration and composition settings, evidence targets, and prompting
protocols. We next analyze these results through fine-grained comparisons across task taxonomy, evidence targets, and prompting effects.

\section{Discussion}

\subsection{Structured Target Construction Is a Persistent Bottleneck}
\label{sec:static-failure-landscape}

The \vnum taxonomy characterizes video-grounded numerical reasoning according
to the computational requirements needed to obtain the final answer. Based on
this taxonomy, we examine whether certain computational requirements
consistently introduce larger errors across models. Specifically, we measure
whether structured target construction presents a recurring cross-model
challenge compared with direct enumeration and quantitative composition. Exact sign tests use the model
as the unit of analysis, with Holm correction across the six reported task and
evidence contrasts. For each model, we define the CSE contrast as

\[
d_{\mathrm{CSE}} =
\mathrm{Acc}_{\mathrm{CSE}} -
\tfrac{1}{2}\bigl(
\mathrm{Acc}_{\mathrm{DDE}} + \mathrm{Acc}_{\mathrm{CQR}}\bigr).
\]

Using the end-to-end accuracies in
Table~\ref{tab:vidnum-detailed-results}, the contrast is negative for 23 of
25 evaluated models under NoCoT, with a mean of $-5.52$ percentage points
(two-sided exact sign test, $p_{\mathrm{Holm}}<0.001$). Under CoT, the
contrast remains negative for 22 of 25 models, with a mean of $-5.42$ points
($p_{\mathrm{Holm}}<0.001$). This suggests that constructing the appropriate
quantitative target from video evidence remains challenging, requiring models
to resolve relevant entities, actions, or events under attributes, relations,
temporal boundaries, or aggregation rules.

\subsection{Action-Grounded Composition Reveals a Cross-Model Failure Pattern}

\begin{figure}[t]
  \centering
  \includegraphics[width=\columnwidth]{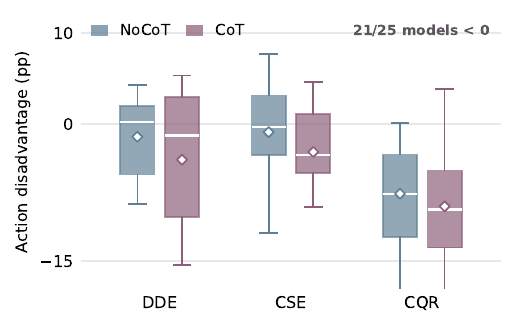}
  \caption{Action disadvantage across \vnum task groups.}
  \label{fig:action-cqr-disadvantage}
\end{figure}

Beyond task-group differences, \vnum also enables analysis of interactions
between the source of visual evidence and the required quantitative
computation. We therefore compare action-grounded questions with object- and
event-grounded questions within each task group. For each model, we define the
action disadvantage as the accuracy on action questions minus the average
accuracy on object and event questions, computed under the same task group and
prompt protocol. This measure captures whether action grounding introduces
additional difficulty beyond the underlying quantitative structure.

The resulting pattern is interaction-specific. DDE and CSE do not show a
consistent action disadvantage across models. In CQR, however, 21 of 25 models
achieve lower accuracy on action-grounded questions under both NoCoT and
zero-shot CoT. The mean gaps are $-7.6$ and $-9.0$ percentage points,
respectively, and both contrasts remain significant after Holm correction
($p_{\mathrm{Holm}}<0.01$).


This finding refines the interpretation of aggregate performance. The issue is
not simply that VLMs struggle with counting actions, nor that compositional
reasoning is universally the most difficult setting. Instead, current models
are particularly challenged when quantities must be derived from dynamic action
evidence and subsequently composed. To check that this pattern, together with
the CSE deficit above, is not solely an artifact of observable benchmark
composition, The technical appendix reports controlled logistic contrasts and
source-video-matched analyses. The CSE deficit and Action-CQR disadvantage remain negative after standardizing over
model identity, video category, clip duration, edited-shot structure, answer
option, and source-video-level variation.

\subsection{Zero-Shot CoT Redistributes Numerical Failures Rather Than Providing a Uniform Gain}
\label{sec:cot-redistribution}

Zero-shot CoT prompting is commonly used to improve reasoning in multimodal
models~\cite{kojima2022large}. However, aggregate accuracy changes do not
reveal whether the same types of video-grounded numerical problems are
improved. \vnum enables this analysis because each question is annotated by
task structure and evidence target. We therefore compare paired NoCoT and CoT
predictions for every question. For each model and category, a question can
remain correct, remain incorrect, be \emph{recovered} by CoT, or be
\emph{broken} by CoT. To separate performance changes from output-format
compliance, we include only questions for which both responses can be parsed;
the thin gray strips in Figure~\ref{fig:cot-transition-groups} separately
report end-to-end CoT parsing failures.

Figure~\ref{fig:cot-transition-groups} shows that CoT does not provide a
uniform improvement across \vnum task groups. Instead, it changes item-level
correctness in both directions, with recovered and broken cases varying across
models and categories. A single accuracy difference is therefore insufficient
to characterize the effect of CoT: even when the net change is near zero, many
questions may be recovered or disrupted. CoT changes the distribution of
problems that models solve rather than uniformly improving numerical reasoning.

The effect of CoT is also dependent on task structure and evidence target.
Across evaluated open-weight models, macro-averaged changes are nearly neutral
for DDE and CQR ($+0.1$ percentage points each), but negative for CSE
($-1.8$ points). When grouped by evidence target, action questions show the
largest average decrease ($-2.5$ points), followed by event questions
($-1.2$ points), while object questions show a small gain ($+0.8$ points).
These trends are not universal across models. The proprietary models exhibit
different patterns: the Gemini pair and GPT-5.5 both show positive changes on
CSE, while GPT-5.5 remains nearly neutral across evidence targets.

These results demonstrate the diagnostic value of \vnum: by jointly characterizing task structure and evidence targets, the benchmark moves beyond aggregate model ranking to reveal persistent computational bottlenecks, evidence-specific failure patterns, and the effects of reasoning interventions such as CoT.


\section{Limitations}

\vnum diagnoses observable final-answer behavior rather than the internal
mechanisms underlying model predictions. Therefore, the identified category-
level failure patterns should be interpreted as behavioral evidence rather than
mechanistic explanations. In addition, \vnum evaluates video-grounded numerical reasoning through curated
video-question pairs and may not capture all forms of interactive, streaming,
or open-ended video understanding. The benchmark is also constructed from
publicly available online videos, and its coverage reflects the domains and
visual contexts available during collection.

\section{Data Release and License}

We will release the \vnum question files, annotations, evaluation scripts, and
answer parsing tools for research use. The annotations and question files will
be released under the CC BY 4.0 license, and the code will be released under
the MIT License. \textbf{We do not redistribute, re-license, or claim ownership of the
original online videos; these videos remain subject to the terms and
copyrights of their respective platforms and content owners.} The release
provides video identifiers, source information, and timestamps to support
reproducible evaluation while respecting third-party video ownership.

\section{Conclusion}

We introduced \vnum, a diagnostic benchmark for video-grounded numerical
reasoning. Through a computation-oriented taxonomy and orthogonal evidence
annotations, \vnum reveals that current VLM failures are structured rather than
uniform. Our analyses show that structured target construction remains a
persistent cross-model bottleneck, action-grounded quantitative composition
forms a recurring failure pattern, and zero-shot CoT changes item-level
correctness without providing consistent improvements.

Beyond measuring overall accuracy, \vnum provides a framework for identifying
where numerical reasoning fails and for evaluating how future models improve
their construction, grounding, and composition of quantities from dynamic
video evidence.
\bibliography{aaai2027}


\end{document}